\documentclass[preprint,3p,12pt,times,sort&compress]{elsarticle}

\usepackage{amssymb}
\usepackage{amsmath}

\usepackage{times}
\usepackage{soul}
\usepackage[small]{caption}
\usepackage{xcolor}
\usepackage{amssymb}
\usepackage{multirow}
\usepackage{colortbl}
\usepackage{subfigure}
\usepackage{tabularray}
\usepackage{arydshln}
\usepackage{comment}
\usepackage{bm}
\usepackage{caption}
\usepackage{bicaption}
\usepackage{hyperref}
\usepackage{bbding}
\usepackage{pifont}
\usepackage[switch]{lineno}
\usepackage[figuresright]{rotating}
\usepackage{wrapfig}

\newcommand{\cmark}{\ding{51}}%
\newcommand{\xmark}{\ding{55}}%

\definecolor{mycolor}{rgb}{0,0,1}
\definecolor{kanbudongcolor}{rgb}{0,1,1}
\definecolor{weijia_color}{RGB}{255,92,51}
\newcommand{\e}{\text{e}}
\allowdisplaybreaks[2]

\makeatletter
\newcommand\figcaption{\def\@captype{figure}\caption}
\newcommand\tabcaption{\def\@captype{table}\caption}
\makeatother

\journal{Expert Systems with Applications}

\begin{document}

\begin{frontmatter}



\title{EPL: Empirical Prototype Learning for Deep Face Recognition}


\author[BigData,CVI]{Weijia Fan}
\author[CVI]{Jiajun Wen}
\author[BigData,CVI,Birmingham]{Xi Jia}
\author[BigData,CVI]{Linlin Shen}
\author[BigData,CVI,Aqara]{Jiancan Zhou}
\author[BigData,CVI]{Qiufu Li\corref{cor1}}

\affiliation[BigData]{organization={National Engineering Laboratory for Big Data System Computing Technology},
            addressline={Shenzhen University},
            city={Shenzhen},
            postcode={518060},
            country={China}}

\affiliation[CVI]{organization={Computer Vision Institute},
            addressline={Shenzhen University},
            city={Shenzhen},
            postcode={518060},
            country={China}}

\affiliation[Birmingham]{organization={School of Computer Science},
            addressline={University of Birmingham},
            city={Birmingham},
            country={UK}}

\affiliation[Aqara]{organization={Aqara},
            addressline={Lumi United Technology Coopration},
            city={Shenzhen},
            country={China}}

\cortext[cor1]{Corresponding author}

\begin{abstract}
    Prototype learning is widely used in face recognition, which takes the row vectors of coefficient matrix in the last linear layer of the feature extraction model as the prototypes for each class. When the prototypes are updated using the facial sample feature gradients in the model training, they are prone to being pulled away from the class center by the hard samples, resulting in decreased overall model performance. 
    In this paper, we explicitly define prototypes as the expectations of sample features in each class and design the empirical prototypes using the existing samples in the dataset. We then devise a strategy to adaptively update these empirical prototypes during the model training based on the similarity between the sample features and the empirical prototypes. Furthermore, we propose an empirical prototype learning (EPL) method, which utilizes an adaptive margin parameter with respect to sample features. EPL assigns larger margins to the normal samples and smaller margins to the hard samples, allowing the learned empirical prototypes to better reflect the class center dominated by the normal samples and finally pull the hard samples towards the empirical prototypes through the learning.
    The extensive experiments on MFR, IJB-C, LFW, CFP-FP, AgeDB, and MegaFace demonstrate the effectiveness of EPL. Our code is available at \href{https://github.com/WakingHours-GitHub/EPL}{https://github.com/WakingHours-GitHub/EPL}.

\end{abstract}




\end{frontmatter}


\section{Introduction}
Face Recognition (FR)~\cite{deng_arcface_2022,wang_cosface_2018,li_2017_integrated,zhou_uniface_2023,li_2023_unitsface} is drawing considerable attention in both the academic and industrial communities,
which includes $1$:$1$ face verification and $1$:$n$ face identification. 
While the $1$:$1$ face verification aims to determine whether two given facial samples belong to the same individual by comparing their feature similarity with a pre-defined threshold,
the $1$:$n$ face identification involves computing the feature similarities between a newly presented facial sample and $n$ pre-registered standard facial samples, and the facial sample is then categorized into the class of the standard sample with the maximum similarity. 
Therefore, both of these two face recognition tasks aim for high similarity among samples from the same individual and low similarity among samples from different individuals. 
With the requirements of face recognition, it has developed metric learning based methods and prototype learning based methods.

Metric learning-based methods explore the distance information of the positive sample pairs (sample pair from the same class) and that of negative sample pairs (sample pair from different classes) to train feature extraction models, aiming for high positive sample-to-sample similarities and low negative ones, such as Triplet Loss~\cite{hoffer_deep_2018, Yang_2018_triplet_center}, Contrastive Loss~\cite{oord_representation_2019}.
These methods either combine the positive and negative sample pairs before the model training or dynamically combine sample pairs online during the training. 
Both these two modes would exponentially increase the data size, compelling the researchers to optimize the feature extraction models using only a subset of the sample pairs, due to the constraints on storage space or training time. 
Therefore, the pure metric learning based methods cannot exploit the whole metric information that is specific to each sample, which results in relatively weak performance. 

The prototype learning-based method in face recognition is derived from classification, such as NormFace~\cite{wang_normface_2017}, CosFace~\cite{wang_cosface_2018} and ArcFace~\cite{deng_arcface_2022}.
The prototype, also referred to as a class proxy or class center, is considered to encapsulate all the sample information belonging to its respective class. 
Typical prototype learning based methods usually employ the SoftMax loss function to train the model. 
For samples from class $i$, these methods minimize the distance between their feature vectors and the $i$-th row vector (i.e., $W_i$) in the coefficient matrix of the model's last linear layer, while simultaneously maximizing the distances from the other row vectors. Consequently, the row vectors are trained to serve as class centers or prototypes in these methods. 
However, 
these prototypes are learnable parameters, updated using sample gradients during the model training, making them susceptible to being ``pulled away'' by the hard samples due to the need to satisfy the objective of minimizing the loss value.
Therefore, these prototypes may not accurately represent the true class centers,
leading to a degradation in the models' ability to capture the intrinsic characteristics of each class, ultimately decreasing the overall performance.
There are hybrid methods combining the advantages of metric learning and prototype learning, such as MixFace~\cite{jung_mixface_2022}, Circle Loss~\cite{sun_circle_2020}, VPL~\cite{deng_variational_2021} and UNPG~\cite{jung_unified_2022}. However, these methods have not resolved the influence of hard samples pulling the prototype learning astray.

In this paper, we define prototypes as the expectations of the sample features for each class and design empirical prototypes using the features of existing samples. The empirical prototypes depict the centroid of the sample features, where the features from the easily learnable samples contribute more to them. 
Consequently, the empirical prototypes can more effectively reduce the distance between the hard samples and the corresponding class center. 
During the training, we directly use the sample features to update the empirical prototypes instead of the sample gradients. 
In addition, in the updating, we introduce adaptive parameters to dynamically adjust the updating weight based on the similarity between samples and empirical prototypes. Furthermore, we propose an empirical prototype learning (EPL) method for face recognition, incorporating adaptive margin parameters to enlarge the distance between sample features with the other class centers. In summary, the main contributions of this paper are as follows:
\begin{itemize}
    \item We explicitly define the prototype as the expectations of sample feature for each class and propose the empirical prototype (EP) for deep feature learning.
    \item We design an adaptive updating strategy for the empirical prototypes, automatically adjusting the updating weights according to the similarity between sample features and their corresponding empirical prototypes.
    \item We propose empirical prototype learning (EPL) for face recognition, integrating with adaptive margin, which can be combined with the existing prototype learning methods, and consistently enhance their performance.
    \item We demonstrate the superior performance of the proposed EPL, by conducting extensive experiments on the popular benchmark datasets for face verification and face identification, including MRF, IJB-C, LFW, CFP-FP, AgeDB, and MegaFace. 
\end{itemize}

\section{Related Work}
Face recognition (FR) tasks aim to distinguish different identities via learning discriminative representation. There are two kinds of commonly employed methods for learning representations, namely metric learning and prototype learning.

\subsection{Metric Learning}
Metric learning contrast positive and negative pairs of facial samples to extract meaningful facial representations at the sample-wise level, which leverages the assumption that similar samples should be closer together and dissimilar ones should be farther apart in a well learned representation space.

Recent works employ metric loss achieves remarkable performance, such as unsupervised leaning~\cite{yin_real-time_2023,he_momentum_2020,wu_unsupervised_2018,chen_simple_2020}, which benefit from data augmentation~\cite{zhang_2018_mixup}, powerful encoders~\cite{vincent_2008_extracting}, pretext tasks~\cite{wu_2018_unsupervised}, and the formulation of metric learning loss functions~\cite{nce_2010,oord_representation_2019}. 
In the face recognition, the success of metric loss can be attributed to two key components: 
(1) the improvement of pair-wise loss functions \cite{sun_deep_2014-1,schroff_facenet_2015,hoffer_deep_2018} and (2) the combination with prototype learning.

DeepID2~\cite{sun_deep_2014-1} combines face identification loss and verification loss to reduce intra-class variations and enlarge inter-class differences.
FaceNet~\cite{schroff_facenet_2015} designs triplet loss \cite{hoffer_deep_2018} using three-element tuples, to optimize model by minimizing the distance between the anchor and positive sample and maximizing the distance between the anchor and negative sample.
Circle loss \cite{sun_circle_2020} adopts a pair similarity re-weighting mechanism to highlight the less-optimized pair similarities, using a unified formula and flexible optimization approach in the deep feature learning.
MagFace~\cite{meng_magface_2021} integrates feature magnitude to balance intra-class variations, enhancing face recognition by ensuring higher quality features are closer to the class center while maintaining a structured distribution within classes.
AdaFace~\cite{kim_adaface_2022} proposed a loss function that utilizes an adaptive margin based on image quality, approximating image quality with feature norms to differentially emphasize samples according to their difficulty, thereby improving face recognition performance across varying image qualities.
About mixed resolution face recognition, \cite{ullah_2024_low_resolution} enhances face recognition performance on low-resolution images by introducing a degradation model to synthesize more realistic low-resolution images from high-resolution ones and utilizing an attention-guided distillation approach, where attention maps guide the feature transfer from a high-resolution teacher network to a low-resolution student network, improving robustness across different resolutions. Xiaojin et al.~\cite{fan_2023_joint} proposed the Joint Coupled Representation and Homogeneous Reconstruction (JCRHR) method for multi-resolution small sample face recognition, which integrates an analysis dictionary with a synthetic dictionary for coupled representation learning, enhances coherence among coding coefficients at different resolutions.

MixFace~\cite{jung_mixface_2022} combines the metric learning and prototype learning, by directly adding the classification loss and the metric loss. 
UNPG~\cite{jung_unified_2022} combines metric and classification pair generation strategies into a unified approach to alleviate the mismatch between the sampled pairs and all negative pairs, enhancing face recognition models' discriminative feature space.
The recent UniTSFace~\cite{li_2023_unitsface} designs a unified threshold integrated sample-to-sample loss to separate the all positive and negative facial sample pairs.

However, the metric learning based methods cannot fully utilize the metric information of all sample pairs due to the extremely large number of sample pairs, resulting in relatively weak face recognition performance.

\subsection{Prototype Learning}
The prototype learning maximizes the similarity between the sample and its corresponding category prototype, and minimizes that between the sample and other prototypes.

The early works, such as DeepFace~\cite{taigman_deepface_2014} and VGGFace~\cite{cao_vggface2_2018}, 
employ a classification loss, i.e., Softmax loss, to optimize similarities between the sample features and the prototypes. 
However, due to that FR is a typical fine-grain task, it requires a more compact feature space, which cannot be well learned by only the vanilla Softmax loss. 
The normalized Softmax loss~\cite{wang_normface_2017,L2_constrained_2017} and marginal Softmax loss~\cite{wang_cosface_2018,deng_arcface_2022,zhou_uniface_2023,meng_magface_2021} are introduced in FR to alleviate the above issue.


CosFace~\cite{wang_cosface_2018} introduces an additive cosine margin to the original softmax loss, which simultaneously enhances the intra-class compactness and inter-class discrepancy by redefining the decision boundaries with cosine values and enforcing precise learnable angular margins between different classes.
ArcFace~\cite{deng_arcface_2022} directly adds an additive angular margin penalty between cosine scores and ground truth target weights in the geodesic distance of hypersphere manifold, which simultaneously enhances intra-class compactness and inter-class discrepancy.
Variational Prototype Learning (VPL)~\cite{deng_variational_2021} addresses the limitations of prototype learning by treating class prototypes as distributions rather than fixed points in feature space. This approach allows for a dynamic comparison between sample-to-prototype and sample-to-sample, thereby encouraging the stochastic gradient descent (SGD) solver to adopt a more exploratory strategy, which enhances performance.

Recently, the research of unified threshold drawing attention~\cite{li_2024_rediscovering,zhou_uniface_2023,li_2023_unitsface}, UniFace~\cite{zhou_uniface_2023} introduced a unified threshold between sample and prototype, and proposed a Unified Cross Entropy Loss function (UCE Loss) in face recognition. The unified threshold satisfies the requirement that the between the smallest positive sample-to-class similarity is greater than the similarity between the largest negative sample-to-class similarity. X2-Softmax~\cite{xu_2024_x2softmax} introduces adaptive angular margins that increase with angle between classes, providing a more intuitive and flexible approach to enhance feature separability. 

However, prototype learning is easily influenced by hard samples, which prevents it from truly and accurately representing the class center, thereby compromising model accuracy and generalization.

\begin{figure*}[ht]
    \centering
    \includegraphics[scale=0.65]{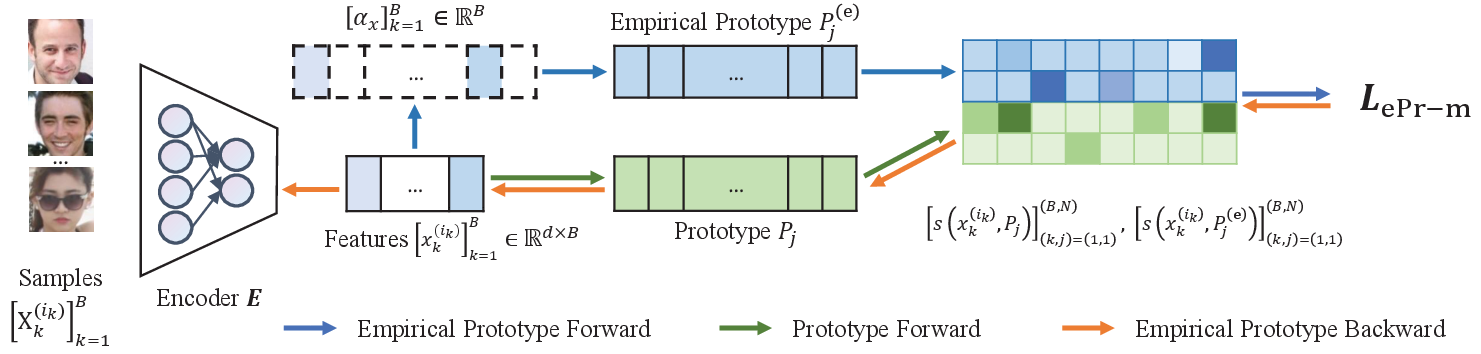}
    \caption{The pipeline of Empirical Prototype Learning (EPL) for deep face recognition. For a batch containing $B$ samples, $[X_k^{(i_k)}]_{k=1}^B$, the encoder extract the features, $[x_k^{(i_k)}]_{k=1}^B\in\mathbb R^{d\times B}$. Then, the adaptive coefficients $\alpha_x$ are calculated to update the empirical prototype corresponding to the current mini-batch of samples $P^{(\text{e})}_{i_k}$. Finally, the similarities, $s(x_k^{i_k}, P_j)$ and $s(x_k^{i_k}, P_j^{\text{(e)}})$, are used to calculate the loss $L_{\text{ePr-m}}$ for the encoder training.}
    \label{fig:method_framework}
\end{figure*}

\section{Method}
In this section, we first revisit the commonly used prototype learning and analyze its drawbacks. Then, we design the empirical prototypes and the empirical prototype learning (EPL) method.

\subsection{Preliminary}
Suppose $\bm E$ is a face encoder trained on a sample set
\begin{align}
    D=\bigcup_{i=1}^N D_i,
\end{align} 
which consists of facial samples captured from $N$ identities, 
and $D_i$ denotes the subset containing the facial sample of the same identity $i$.
For any sample $X\in D$, one can generate its feature vector $x = \bm E(X) \in\mathbb R^d$ using the encoder, resulting in a feature set 
\begin{align}
    \label{eq:feature_set}
    F = \bigcup_{i=1}^N F_i = \bigcup_{i=1}^N\Big\{x^{(i)} = \bm E(X^{(i)})| X^{(i)}\in D_i\Big\},
\end{align}
where $X^{(i)}$ denotes facial sample captured from identity $i$, 
and $x^{(i)}$ is the feature extract by $\bm E$.

\subsection{Prototype Learning}
For each identity $i$, its prototype $P_i$ represents the center of its sample features. 
In the Softmax-based methods, the prototype is considered to be stored in the coefficient matrix of the last linear layer, i.e., $P_i = W_i$.
The canonical prototype learning applies Softmax-based loss or its variants,
\begin{align}
    \label{eq:softmax_loss}
    L_{\text{Pro}}(x^{(i)}) &= -\log\frac{\exp(\gamma~s(x^{(i)}, P_i))}{\sum_{j=1}^N\exp(\gamma~s(x^{(i)}, P_j))}\nonumber\\
                            &= \log\Big[1 + \sum_{j=1\atop j\neq i}^N \frac{\exp\big(\gamma~s(x^{(i)}, P_j)\big)}{\exp\big(\gamma~s(x^{(i)}, P_i)\big)}\Big],
\end{align}
where $\gamma$ denotes scale factor and is set to 64 following previous works~\cite{wang_cosface_2018,deng_arcface_2022}. $s(\cdot, \cdot)$ is a similarity function,
which is usually taken as cosine function, i.e.,
\begin{align}
    \label{eq:cosine_function}
    s(x, P) = \frac{\left\langle x, P\right\rangle}{\|x\|\|P\|}, \quad\forall~x,P\in\mathbb R^d.
\end{align}

In the optimization of an encoder, the loss in Eq. (\ref{eq:softmax_loss}) maximizes the similarity between the feature $x^{(i)}$ and $W_i$,
which ensures that the $W_i$ can carry all the feature information of the identity $i$ after the training.
Therefore, $W_i$ is referred to as the class center, proxy, or prototype for identity $i$ in literature~\cite{deng_variational_2021,zhou_uniface_2023}.

To enhance discriminability of inter-class and compactness of intra-class, usually consider a form of marginal loss, 
\begin{align}
    \label{eq:marginal_softmax_loss}
    L_{\text{Pro-m}}(x^{(i)}) &= -\log\frac{\e^{\gamma[s(x^{(i)}, P_i) - m]}}{\sum_{j=1}^N\e^{\gamma s(x^{(i)}, P_j)}}\nonumber\\
                            &= \log\Big(1 + \sum_{j=1 \atop j \neq i}^N \frac{\e^{\gamma s(x^{(i)}, P_j)}}{\e^{\gamma[s(x^{(i)}, P_i) -m] }}\Big),
\end{align}
where $m$ is a margin hyper-parameter. 

In prototype learning, the prototype $P_j$ is updated using its gradient in the backpropagation,
\begin{align}\label{eq_pl_updating}
    P_j = P_j - \varepsilon \frac{\partial L_{\text{Pro}}(x^{(i)})}{\partial P_j}, \quad \forall j,
\end{align}
where $\varepsilon$ is the learning rate, and 
\begin{align}
    \label{eq_pl_gradient_positive}
    \frac{\partial L_{\text{Pro}}(x^{(i)})}{\partial P_i} &=\Big(\frac{\e^{\gamma [s(x^{(i)}, P_i) - m] }}{\sum_{k=1}^{N} \e^{\gamma s(x^{(i)}, P_k)}}-1\Big)\cdot x^{(i)},\\
    \label{eq_pl_gradient_negative}
    \frac{\partial L_{\text{Pro}}(x^{(i)})}{\partial P_{j}} &=\frac{\e^{\gamma s(x^{(i)}, P_j)}}{\sum_{k=1}^{N} \e^{\gamma s(x^{(i)}, P_k)}}\cdot x^{(i)},~j \neq i.
\end{align}
Due to hard samples, the updating strategy for prototype learning has two limitations.
Suppose that $X^{(i)}_{\text{har}}$ and $X^{(i)}_{\text{nor}}$ are respectively hard sample and normal sample from identity $i$.
\begin{itemize}
    \item L-A: Combining Eqs. (\ref{eq_pl_updating}) and (\ref{eq_pl_gradient_positive}), 
the prototype $P_i$ tends to be pulled towards to the hard sample's feature $x^{(i)}_{\text{har}} = \bm E(X^{(i)}_{\text{har}})$, as the updating coefficient corresponding to the hard sample is greater than that of the normal sample, resulted from the lower hard sample similarity $s(x^{(i)}_{\text{har}}, P_i)$ and higher normal one $s(x^{(i)}_{\text{nor}}, P_i)$.
The prototypes learned in this strategy will be pulled away from the class center by the hard samples.

\item L-B: For the hard sample feature $x^{(i)}_{\text{har}}$ from identify $i$, there exists a prototype $P_{j'}, j'\neq i$, 
such that $s(x^{(i)}_{\text{har}},P_{j'}) \geq s(x^{(i)}_{\text{har}},P_{i})$, 
and the negative similarity $s(x^{(i)}_{\text{har}},P_{j'})$ would be even larger than positive ones $s(x^{(j')}_{\text{nor}},P_{j'})$ of some samples from identity $j'$.
Then, according to Eqs. (\ref{eq_pl_updating}) and (\ref{eq_pl_gradient_negative}), 
the prototype $P_{j'}$ would be pushed more severely by the negative hard sample's feature $x^{(i)}_{\text{har}}$
than pulled by the positive normal sample's feature $x^{(j')}_{\text{nor}}$.
This drawback would decrease the representation of prototype $P_{j'}$ for the identity $j'$. 
\end{itemize}

\subsection{Empirical Prototype}

In this paper, we define the prototype of identity $i$ as the expectation of the features of this identity, i.e.,
\begin{align}
    \label{eq:prototype_define}
    P_i := \mathbb{E}(x^{(i)}),\quad\forall~i,
\end{align}
where $\mathbb{E}$ denotes the expectation operator. 

Clearly, in practice, it is not possible to obtain this expectation precisely; 
however, one can estimate the expectation using the features of all samples captured from the identity, 
\begin{align}
    \label{eq:prototype_empirical}
    P_i^{(\text{e})} := \hat{\mathbb{E}}(\{x^{(i)}\}) = \hat{\mathbb{E}}(F_i),
\end{align}
where $\hat{\mathbb{E}}$ is a expectation estimation operator, 
and $F_i$ is the feature set for identity $i$.
We call the estimated expectation as \textbf{empirical prototype}. 

In practice, one can flexibly design its prototype estimation operator and the implementations based on its task requirements. 
For better feature representation, it is essential to design the prototype estimation operator in conjunction with the optimization process of the model 
and continually update the estimated prototypes during model iterations. 
In VPL~\cite{deng_variational_2021}, the authors consider $W_i$ as the prototype center and update the prototype using back-propagation based on the variational information of sample features during the model optimization process. 

In this paper, we estimate prototypes in the following ways:
(1) Directly generate and update prototypes using the sample features. 
This enables the constructed empirical prototype to be seamlessly integrated with existing methods without modifying them. 
In addition, the prototypes can be rapidly updated in the forward process of the model without the secondary updates through back-propagation.
(2) Adaptively update the prototypes. In the early stages of training, it is important to retain more sample feature information to facilitate quicker learning of prototype information related to identity. 
In the later stages of training, preserving more information about the prototypes themselves is crucial to stabilize the optimization process.
According to these requirements, we generate and updata prototypes following the below steps:

\begin{enumerate}
    \item Before training of the encoder $\bm E$, we randomly generate a initial prototype $P^{(\text{e})}_i$ for each identity $i$;
    \item During the training, given any sample $X \in D$, if it is captured from identity $i$, 
            the empirical prototype $P^{(\text{e})}_i$ is updated using the feature $x = \bm E(X)$,
    \begin{align}
        \label{eq:prototype_updating}
        P^{(\text{e})}_i = \alpha_x \cdot P^{(\text{e})}_i + (1- \alpha_x)\cdot x, 
    \end{align}
    where
    \begin{align}
        \label{eq:alpha_x}
        \alpha_x = \sigma(s(x, P^{(\text{e})}_i))
    \end{align}
    is an adaptive updating coefficient generated using the feature $x$ and its prototype $P_i^{(\e)}$,
    and ``$\sigma$'' is an activation function to adjust the updating coefficient into an appropriate range.
\end{enumerate}

Clearly, in Eq. (\ref{eq:prototype_updating}), the empirical prototype $P_i^{\text{(e)}}$ is updated only using its positive samples' features during the forward propagation, 
without being biased by hard samples from the other identities, thus avoiding the limitation ``L-B'' in the prototype learning. 

Additionally, the updating strategy defined in Eq. (\ref{eq:prototype_updating}) balances the strength of empirical prototype updates at different training stages. In the early stages of training, when the encoder $\bm E$ has not yet learned well sample features, the empirical prototypes computed by the sample features cannot well reflect the expectation of sample features, resulting in low similarity between sample features and their empirical prototypes. Therefore, at this stage, the updating strategy in Eq. (\ref{eq:prototype_updating}) adopt a larger update coefficient of $1-\alpha_x$ to update the empirical prototype using newly obtained sample features, while retaining fewer historical features in the empirical prototype. 
In contrast, in the terminal stage of training, when the encoder has learned to extract well features, the empirical prototype can better reflect the expectation of the sample features. The similarity between sample feature and the empirical prototype is high, and a smaller update coefficient is used to update the empirical prototype, preserving more historical features in the empirical prototype.

\subsection{Empirical Prototype Learning}

Similar to the prototype learning loss in Eq. (\ref{eq:softmax_loss}), we define the empirical prototype learning loss as 
\begin{align}
    \label{eq:empirical_prototype_learning_loss}
     L_{\text{ePr}}(x^{(i)}) = &-\log\frac{\e^{\frac{1}{\tau} s(x^{(i)}, P^{(\e)}_i)}}{\sum_{j=1}^N\e^{\frac{1}{\tau} s(x^{(i)}, P^{(\e)}_j) }} \nonumber\\
    = &\log\Big[1 + \sum_{j=1\atop j\neq i}^N\frac{\e^{\frac{1}{\tau} s(x^{(i)}, P^{(\e)}_j) }}{\e^{\frac{1}{\tau} s(x^{(i)}, P^{(\e)}_i)}} \Big],
\end{align}
where $\tau$ is an inverse temperature factor, playing the same role as $\gamma$ in Eq.~(\ref{eq:softmax_loss}), thus we set it to 1/64.

To comprehensively leverage the capabilities of both prototype learning and empirical prototype learning,
we combine them together and introduce distinct margins,
\begin{align}
    \label{eq:empirical_prototype_prototype}
    L_{\text{ePr-m}}(x^{(i)}) = 
    \log&\bigg[1+\sum_{j=1\atop j\neq i}^N \frac{\e^{\frac{1}{\tau}s(x^{(i)}, P_j^{(\e)}) }}{\e^{\frac{1}{\tau}s(x^{(i)}, P_i^{(\e)}) - \beta m_{x^{(i)}} }  }
     + \sum_{j=1\atop j\neq i}^N \frac{\e^{\gamma s(x^{(i)}, P_j) }}
                            {\e^{\gamma[s(x^{(i)}, P_i) -m]} }\bigg],
\end{align}
where 
\begin{align}
    \label{eq:adaptive_margin}
    m_{x^{(i)}} = \text{Detach}\Big(\frac{1}{\tau} s(x^{(i)}, P_i^{(\e)})\Big)
\end{align}
is an adaptive margin for the empirical prototype learning, and $\beta$ is a scalar factor. 

The adaptive margin defined by Eq. (\ref{eq:adaptive_margin}) only utilizes the value of positive similarity $s(x^{(i)}, P_i^{(\e)})$ and does not participate in the calculating of gradients with respect to the sample feature $x^{(i)}$ during the back propagation. In addition, when training the model using loss $L_{\text{ePr-m}}(x^{(i)})$, the empirical prototypes are only updated using Eq. (\ref{eq:prototype_updating}) during the forward propagation and are not updated during the back propagation.

In $L_{\text{ePr-m}}(x^{(i)})$, the prototype term adopts a fixed margin, inspired by references \cite{wang_cosface_2018,deng_arcface_2022}. The adaptive margin has been also utilized for the prototype learning \cite{kim_adaface_2022,meng_magface_2021}, while larger margins are assigned to the hard samples and smaller margins to the normal samples. 
In contrast, according to the similarities, $L_{\text{ePr-m}}(x^{(i)})$ assigns smaller margin to the hard samples and larger margins to normal samples with respect to the empirical prototypes, which makes the encoder to focus more on learning from the majority of normal samples in the dataset during its training, and helps to learn better empirical prototypes. This adaptive margin strategy suppresses the influence of the hard samples, preventing them from pulling the empirical prototypes away from the sample feature centers.

To extract more discriminative sample features, the previous Softmax loss based methods with adaptive margins, such as AdaFace and MagFace, assign larger margins to the hard samples in the model training. Our EPL, in contrast, first constructs more discriminative class-level empirical prototypes using the features of normal sample, which are predominant in the dataset, and then pulls the hard samples towards the class center via these empirical prototypes. Therefore, EPL assigns larger margins to the normal samples and smaller margins to the hard samples during the model training.

Fig.~\ref{fig:method_framework} presents the pipeline of the proposed method.
In every iteration of the training, the encoder $\bm E$ is fed with a batch of samples, $[X_k^{(i_k)}]_{k=1}^B$,
which are transformed into their features $[x_k^{(i_k)}]_{k=1}^B\in\mathbb R^{d\times B}$.
The features are first applied to update the empirical prototypes, $P_i^{(\e)}$, 
according to their labels $[i_k]_{k=1}^B$ and Eqs. (\ref{eq:prototype_updating}) and (\ref{eq:alpha_x}).
Then, we compute their positive and negative similarities with the prototypes and the empirical prototypes, i.e.,
\begin{align}
    \label{eq:positive_negative_similarities}
    \Big[s(x_k^{(i_k)}, P_j)\Big]_{(k,j)=(1,1)}^{(B,N)},~\Big[s(x_k^{(i_k)}, P_j^{(\e)})\Big]_{(k,j)=(1,1)}^{(B,N)}.
\end{align}

Finally, we calculate the loss for each feature $x^{(i_k)}_k$ using Eq. (\ref{eq:empirical_prototype_prototype}),
and update the parameters of encoder $\bm E$ and the prototype $P_i$ with the back-propagation.

\section{Experiments and Results}
In the experiments, we employ ResNets \cite{li_2023_unitsface, Partial_Birds_2022_CVPR} as facial feature extraction models to validate the effectiveness of the proposed EPL.

\subsection{Training Datasets and Details}

\paragraph{Datasets}
As Table \ref{tab:settings} shows, we train the models on four popular facial datasets, including CASIA-WebFace~\cite{yi_learning_2014} (consisting of 0.5 million images of 10K identities), Glint360K~\cite{Xiang_2021_PartialFC} (comprising 17.1 million images of 360K identities), WebFace4M (with 0.2 million identities and 4.2 million face images), and WebFace12M (with 0.6 million identities and 12.7 million face images), where WebFace4M and WebFace12M are subset of WebFace42M~\cite{zhu_webface260m_2021}. 

\paragraph{Preprocessing}
In the preprocessing stage, face images were aligned and resized to the size of $112 \times 112$, then their three channels were normalized using the mean of (0.5, 0.5, 0.5) and standard deviation of (0.5, 0.5, 0.5). For data augmentation, a horizontal flip was applied with a 50\% of chance. 

\begin{wraptable}{r}{0.52\textwidth}
	\centering
    \caption{The training datasets and details.}
    \label{tab:settings}
    \centering
    \footnotesize
    \begin{tabular}{c!{\color{black}\vrule}c!{\color{black}\vrule}c!{\color{black}\vrule}c} 
    \hline
    Dataset                        & Description & \multicolumn{2}{c}{Training settings}\\\hline
                                   &     & Epoch   & 28       \\
    CASIA-                  &  0.5M im.   & Batchsize     & 512 \\
    WebFace0.5M                         &  10K ID       & Initial  lr & 0.1 \\
                                   &          & Schedule  & Step \\ 
    \hline
    \multirow{2}{*}{WebFace4M}        & 4.2M im.                    &   &   \\ 
                                    & 0.2M ID                       & Epoch  & 20 \\
    \cline{1-2}
    \multirow{2}{*}{WebFace12M}     & 12.7M im.                   &  Batchsize  & 1024 \\ 
                                &0.6M ID & Initial  lr&0.1\\
    \cline{1-2}
    \multirow{2}{*}{Glint360K}     & 17.1M im.  & Schedule & Polynomial \\
                                   &  360K ID &  &  \\
    \hline
    \end{tabular}
\end{wraptable}

\paragraph{Training}
We employed different training strategies across the four training datasets following~\cite{li_2023_unitsface}. We adopt customized ResNets as our backbone following~\cite{deng_arcface_2022,li_2023_unitsface}.
On CASIA-WebFace, we trained the modes 28 epochs with a batch size of 512, using step schedule;
the initial learning rate was 0.1, which was decayed by multiplying 0.1 at the ${16}^{th}$ and ${24}^{th}$ epochs, respectively. 
On the WebFace4M, WebFace12M, and Glint360K, we adopted the polynomial decay strategy (power=2), the batch size was set to 1024, and the maximum epoch was set to 20. 
All models were trained using SGD optimizer with weight decay of 5e-4 and momentum of 0.9 on eight NVIDIA V100 GPUs. In each training, we start EPL at the fourth epoch. 
While EPL requires only 122M extra memory in the training, and its speed is slightly reduced from 1346 to 1330 im./sec., it does not introduce any additional costs in the evaluation.


\subsection{Evaluation Settings}
In the evaluation, we use the ICCV-2021 Masked Face Recognition Challenge (MFR Ongoing)~\cite{Deng_2021_ICCV}, which contains popular benchmarks such as LFW~\cite{huang_labeled_2008}, CFP-FP~\cite{sengupta_frontal_2016}, AgeDB~\cite{moschoglou_agedb_2017} and IJB-C~\cite{maze_iarpa_2018}, along with its own more challenging subsets, such as the Mask, Children and Globalized Multi-Racial test sets. 
We evaluate the trained models by directly submitting them to the online MFR Ongoing Challenge server and report the corresponding results. 
Specifically, we report the 1:1 verification accuracy on the LFW, CFP-FP and AgeDB datasets. 
In the case of the IJB-C, we report the True Accept Rate (TAR) at False Accept Rate (FAR) levels of 1e-4 and 1e-5. 
For the MFR benchmarks, we report the TARs at FAR=1e-4 for the Mask and Children test sets, and TARs at FAR=1e-6 for the Generalized Multiple Recognition (GMR) test sets.

Besides the ICCV-2021 MFR Challenge, we also evaluate our method on the MegaFace Challenge 1~\cite{megaface_challenge_2016}, which comprises a gallery set with over 1 million images from 690K different identities and a probe set with 3530 images from 530 identities. 
On the Megaface Challenge 1, we report Rank1 accuracy for $1\text{:}n$ identification and TAR at FAR=1e-6 for $1\text{:}1$ verification on `Small' and `Large' protocols.




\begin{figure*}[t]
    \centering
    \begin{minipage}{0.4\textwidth}
        \centering
        \includegraphics[width=\linewidth]{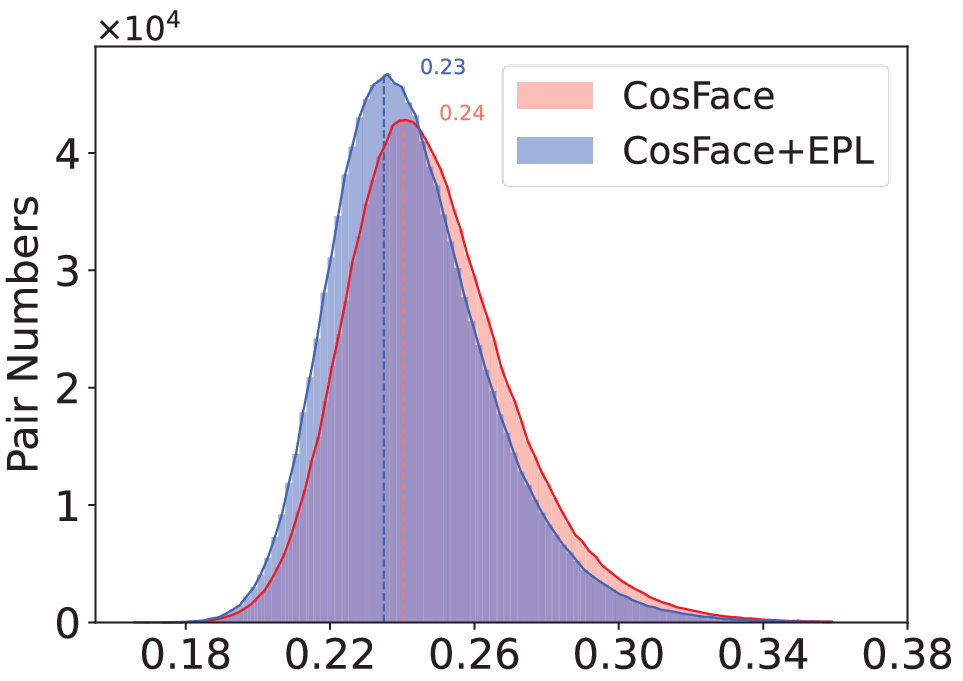}
        \caption{Similarity distribution between samples and different prototypes.}
        \label{fig:impact_of_EPL}
    \end{minipage}%
    \hfill
    \begin{minipage}{0.56\textwidth}
        \centering
        \footnotesize
        \captionof{table}{Performance of prototype learning based methods with and without EPL.}
        \label{tab:epl_effectiveness}
        \setlength{\tabcolsep}{0.6mm}{
            \begin{tabular}{l|c|c|cc|ccc} 
                \hline
                \multirow{2}{*}{Methods} & \multirow{2}{*}{Mask} & \multirow{2}{*}{MR-All} & \multicolumn{2}{c|}{IJB-C}        & \multicolumn{3}{c}{1:1 Veri. Acc.}   \\
                &                       &                         & 1e-5            & 1e-4           & LFW           & CFP           & Age            \\ 
                \hline
                ArcFace     & 38.52           & 42.21           & 9.18             & 48.49           & 99.31           & \textbf{97.07} & 94.51                     \\
                ArcFace+EPL & \textbf{40.90}  & \textbf{48.86}  & \textbf{81.93}   & \textbf{89.29}  & \textbf{99.37}  & 96.47           & \textbf{94.75}            \\ 
                \hline
                CosFace     & 38.79           & 45.12           & 11.30            & 56.65           & 99.36           & \textbf{97.30} & \textbf{94.98}            \\
                CosFace+EPL & \textbf{40.92}  & \textbf{51.92}  & \textbf{83.38}   & \textbf{90.13}  & \textbf{99.45}  & 96.46           & 94.57                     \\ 
                \hline
                UniFace     & 37.30           & 48.17           & 78.99            & 89.18           & 99.45           & \textbf{97.30}  & 94.82                     \\
                UniFace+EPL & \textbf{41.39}  & \textbf{50.26}  & \textbf{82.60}   & \textbf{90.16}  & \textbf{99.55}  & 96.91           & \textbf{95.08}            \\
                \hline
            \end{tabular}
        }
    \end{minipage}
\end{figure*}

\subsection{Ablation Study and Parameter Study} \label{sec:sec_parameter_study}
\paragraph{Empirical prototype learning} 

To demonstrate the effectiveness and compatibility of EPL, we train ResNet50 based feature extraction model using three prototype learning methods including CosFace~\cite{wang_cosface_2018}, ArcFace~\cite{deng_arcface_2022}, and UniFace~\cite{zhou_uniface_2023}, with and without our EPL on the CASIA-WebFace dataset.
Table Table~\ref{tab:epl_effectiveness} shows the results on the MFR Challenge.

One can observe that EPL has consistently improved the overall performance of the three prototype learning methods. When integrating the EPL in CosFace, a significant performance boost from 45.12\% to 51.92\% (+6.8\%) on the MR-All, and from 11.30\% to 83.38\% on IJB-C (1e-5). 
For ArcFace and UniFace, EPL imporoves their performance from 42.21\% and 48.17\% to 48.86\% and 50.26\% on the MR-All, with clear increase of 6.65\% and 2.09\%, respectively.
Furthermore, one can find that EPL also consistently improves the accuracy of the three methods on the hard face samples wearing masks. On the Mask subset, EPL increases the performance of ArcFace, CosFace, and UniFace from 38.52\%, 38.79\%, and 37.30\% to 40.90\%, 40.92\%, and 41.39\%, respectively.
To illustrate that EPL can learn better prototype centers. EPL suppresses the samples from approaching the prototypes of other classes. We computed the negative similarities between each sample and all prototypes of different classes, and Fig.~\ref{fig:impact_of_EPL} shows the distribution of the top 3 maximum negative similarities for each sample. It can be observed that the peak negative similarity of CosFace is 0.24, whereas EPL reduces it to 0.23, indicating that EPL decreases the negative similarities between prototypes and samples from other classes, thus suppressing the prototypes from being pulled by samples of different classes.

\paragraph{Ablation study}
We conduct an ablation study to evaluate each component in EPL. Table~\ref{tab:ablate_experiment} shows the ablation experimental results of ResNet50 with CosFace on CASIA-WebFace. 
The 1:1 verification results on MR-All of the original CosFace is 45.12\%, which is increased to 48.42\% after our EP is integrated into its loss function in the training. When the EP is updated using the proposed strategy or integrated with the adative margin, the verification accuracy is respectively increased to 49.52\% and 49.72\%.
Finally, the optimal performance (51.92\%) is achieved when the complete EPL is integrated into CosFace. 
These results illustrate the effectiveness of each component in our EPL.

\begin{table}[t]
    \centering
    \begin{minipage}{0.48\textwidth}
        \centering
        \caption{The ablation experiment of proposed EPL.}
        \label{tab:ablate_experiment}
        \footnotesize
        \arrayrulecolor{black}
        \begin{tabular}{c|ccc|c} 
        \hline
        \multirow{3}{*}{Method}  &  \multicolumn{3}{c|}{EPL} & \multirow{3}{*}{MR-All}\\ \cline{2-4}                                                         & \multirow{2}{*}{EP}     & Updating & Adaptive  &           \\ 
                                 &                          & Strategy & Margin&\\    \hline
        CosFace                  & $\times$                  & -         & -               & 45.12           \\ 
        \hdashline
        \multirow{4}{*}{\begin{tabular}[c]{@{}c@{}}CosFace \\+EPL\end{tabular}} & $\checkmark$ & $\times$         & $\times$               & 48.42           \\
                 &     $\checkmark$  & $\times$         & $\checkmark$               & 49.76           \\
                 &   $\checkmark$    & $\checkmark$         & $\times$               & 49.52           \\
                 &   $\checkmark$     & $\checkmark$         & $\checkmark$          & \textbf{51.92}  \\
        \hline
        \end{tabular}
    \end{minipage}
    \hfill
    \begin{minipage}{0.48\textwidth}
        \centering
        \caption{Parameter study for $\sigma(\cdot)$ in updating strategy and $\beta$ in adaptive margin.}
        \label{tab:ablation_of_update_and_margin}
        \footnotesize
        \arrayrulecolor{black}
        \begin{tabular}{c|c||c|c} 
        \hline
        $\sigma(\cdot)$   & MR-All       & $\beta$           & MR-All           \\ 
        \hline
        Identity($x$)  & 48.42           & $0.5$          & 46.01           \\
        ReLU($x$)      & 47.87           & $0.6$          & 49.91           \\
        Sigmoid($x$)   & 44.98           & $0.7$          & \textbf{51.92}  \\
        Sigmoid($x-1$) & 46.38           & $0.8$          & 49.17           \\
        Softsign($x$)  & \textbf{49.76}  & $0.9$          & 47.87           \\
        \hline
        \end{tabular}
    \end{minipage}%
\end{table}

\paragraph{Parameter study in updating strategy}
In our updating strategy, an activation function $\sigma(\cdot)$ is adopted to adjust the updating coefficient in Eq.~\ref{eq:alpha_x}. We evaluate the performance of different activate functions, including ReLU~\cite{relu_2011}, Sigmoid~\cite{sigmoid_1991}, Sigmoid($x$-$1$), and Softsign~\cite{Softsign_2009}. Table \ref{tab:ablation_of_update_and_margin} shows the results, and identify indicates updating the empirical prototype directly using the coming sample features in rule of $P_i^{(\text{e})} = x$. 
According to the results in Table~\ref{tab:ablation_of_update_and_margin}, when the Softsign activation function is used in the updating strategy, the accuracy reaches the optimal values of 49.76\% on MR-All. Therefore, we consistently adopt the softsign function in other experiments.

\paragraph{Parameter study in adaptive margin}
We here conduct the parameter study for the scale factor $\beta$ of the adaptive margin in Eq.~\ref{eq:empirical_prototype_prototype}.
Table \ref{tab:ablation_of_update_and_margin} shows the face verification results of ResNet50 trained by CosFace with EPL 
when $\beta$ varies from $0.5$ to $0.9$ with step of $0.1$. When $\beta = 0.7$, the model achieves the best result (51.92\%) on MR-All. Therefore, we fix $\beta = 0.7$ in the other experiments.

\begin{table*}[t]
    \caption{Benchmark results on MFR Ongoing (* denotes re-implement).}
    \footnotesize
    \centering
    \label{tab:mfr_benchmark_results}
    \setlength{\tabcolsep}{0.65mm}{
        \arrayrulecolor{black}
        \begin{tabular}{l|c|ccccccc|cc|ccc} 
        \hline
        \multirow{2}{*}{Method}              & \multirow{2}{*}{\begin{tabular}[c]{@{}c@{}}Network \\Dataset\end{tabular}}      & \multicolumn{7}{c|}{MFR}                                                                                              & \multicolumn{2}{c|}{IJB-C}      & \multicolumn{3}{c}{Verification Acc.}         \\ 
        \cline{3-14}
                                             &                                                                                 & Mask           & Child.         & Afri.          & Cau.           & S-A.           & E-A.           & \textbf{MR-All} & 1e-5           & 1e-4           & LFW            & CFP            & Age             \\ 
        \hline
        Contrastive                          & \multirow{11}{*}{\begin{tabular}[c]{@{}c@{}}ResNet50 \\CASIA0.5M\end{tabular}}  & 6.67           & 10.58          & 12.40          & 18.84          & 13.57          & 10.38          & 12.38           & 46.75          & 58.47          & 95.50          & 74.65          & 82.28           \\
        (N+1)-Tuplet                         &                                                                                 & 26.56          & 28.23          & 28.23          & 50.92          & 47.71          & 24.06          & 38.11           & 73.93          & 83.60          & 99.18          & 96.32          & 94.18           \\
        ArcFace                              &                                                                                 & 38.52          & 31.42          & 45.87          & 63.69          & 59.85          & 7.66           & 42.21           & 9.18           & 48.49          & 99.31          & 97.07          & 94.51           \\
        CosFace                              &                                                                                 & 38.79          & 31.33          & 48.06          & 63.56          & 58.71          & 15.08          & 45.12           & 11.30          & 56.65          & 99.36          & 97.30          & 94.98           \\
        UniFace,$L_{\text{uce-m=0.4}}$*        &                                                                                 & 37.30          & 31.64          & 49.45          & 64.30          & 60.09          & 29.15          & 48.17           & 78.99          & 89.18          & 99.45          & 97.30          & 94.82           \\
        UniFace,$L_{\text{uce-mb-}{\lambda}}$  &                                                                                 & 39.25          & 33.11          & 50.79          & 66.46          & 61.47          & 29.71          & 48.72           & 78.30          & 88.94          & 99.30          & 97.20          & 94.95           \\
        VPL                                  &                                                                                 & 33.86          & 31.39          & 46.52          & 59.93          & 54.07          & 27.18          & 27.18           & 81.38          & 88.44          & 99.30          & 97.07          & 94.75           \\
        AnchorFace                           &                                                                                 & 37.04          & 32.28          & 49.60          & 63.17          & 59.80          & 28.88          & 48.44           & 77.82          & 88.81          & \textbf{99.56} & \textbf{97.48} & \textbf{95.18}  \\
        UNPG                                 &                                                                                 & 38.62          & 33.24          & 49.94          & 63.85          & 59.60          & 29.21          & 48.66           & 77.73          & 88.17          & 99.45          & 97.25          & 94.83           \\
        AdaFace                              &                                                                                 & -              & -              & -              & -              & -              & -              & -               & -              & -              & 99.42          & 96.41          & 94.38           \\
        UniTSFace                            &                                                                                 & 37.98          & 31.73          & 51.45          & 64.89          & 59.73          & 29.56          & 50.28           & 82.64          & 89.84          & 99.41          & 97.35          & 95.13           \\
        EPL                                  &                                                                                 & \textbf{40.92} & \textbf{33.13} & \textbf{51.50} & \textbf{65.95} & \textbf{62.32} & \textbf{31.23} & \textbf{51.92}  & \textbf{83.38} & \textbf{90.13} & 99.45          & 96.46          & 94.47           \\ 
        \hline
        Partial FC                           & \multirow{5}{*}{\begin{tabular}[c]{@{}c@{}}ResNet50\\ WebFace4M\end{tabular}}   & 72.28          & -              & 84.86          & 91.57          & 88.57          & 67.52          & 86.85           & -              & -              & -              & -              & -               \\
        AdaFace                              &                                                                                 & -              & -              & -              & -              & -              & -              & -               & -              & 96.98          & 99.78          & 98.97          & 97.78           \\
        UniFace, $L_{\text{uce-mb-}{\lambda}}$ &                                                                                 & 75.46          & 69.32          & 86.89          & 93.02          & 90.36          & 69.46          & 88.55           & 94.90          & 96.96          & \textbf{99.80} & 98.98          & 97.88           \\
        UniTSFace                            &                                                                                 & 75.93          & 72.00          & 88.17          & 93.68          & 91.40          & 70.55          & 89.65           & \textbf{95.18} & \textbf{97.03} & \textbf{99.80} & \textbf{99.04} & \textbf{97.93}  \\
        EPL                                  &                                                                                 & \textbf{76.01} & \textbf{72.31} & \textbf{88.33} & \textbf{93.47} & \textbf{91.73} & \textbf{71.78} & \textbf{89.76}  & \textbf{95.18} & 97.01          & 99.78          & 98.94          & 97.67           \\ 
        \hline
        Partial FC                           & \multirow{3}{*}{\begin{tabular}[c]{@{}c@{}}ResNet50 \\WebFace12M\end{tabular}}  & 80.08          & -              & 91.14          & 95.00          & 93.61          & 75.55          & 91.82           & -              & -              & -              & -              & -               \\
        UniTSFace *                          &                                                                                 & 81.46          & 79.66          & 92.52          & 95.73          & 94.73          & 77.64          & \textbf{93.14}  & 95.86          & 97.27          & \textbf{99.80} & \textbf{99.27} & \textbf{98.12}  \\
        EPL                                  &                                                                                 & \textbf{82.69} & \textbf{80.10} & \textbf{92.64} & \textbf{95.89} & \textbf{94.77} & \textbf{77.84} & \textbf{93.14}  & \textbf{95.99} & \textbf{97.36} & \textbf{99.80} & 99.01          & 97.93           \\ 
        \hline
        UniTSFace *                          & \multirow{2}{*}{\begin{tabular}[c]{@{}c@{}}ResNet100 \\WebFace12M\end{tabular}} & 86.35          & 86.26          & 95.21          & 97.20          & 96.82          & 82.66          & 95.47           & 96.00          & 97.57          & \textbf{99.80} & \textbf{99.39} & 98.33           \\
        EPL                                  &                                                                                 & \textbf{86.88} & \textbf{88.32} & \textbf{95.81} & \textbf{97.66} & \textbf{97.22} & \textbf{82.89} & \textbf{95.73}  & \textbf{96.43} & \textbf{97.60} & \textbf{99.80} & 99.30          & \textbf{98.37}  \\
        \hline
        \end{tabular}
    }
\end{table*}

\begin{table*}[th]
    \caption{Comparisons on the MegaFace Challenge 1.}
    \label{tab:megaface_result}
    \centering
    \setlength{\tabcolsep}{1.4mm}{
        \arrayrulecolor{black}
        \begin{tabular}{l|llll||l!{\color{black}\vrule}llll} 
        \hline
        Method        & P & R & Iden.           & Veri.           & Method         & P & R & Iden.          & Veri.           \\ 
        \arrayrulecolor{black}\cline{1-1}\arrayrulecolor{black}\cline{2-10}
        Softmax Loss   & S    & \xmark     & 54.85           & 65.92           & ArcFace        & L    & \xmark     & 81.03          & 96.98           \\
        Triplet Loss   & S    & \xmark     & 64.79           & 78.32           & Curricular     & L    & \xmark     & 81.26          & 97.26           \\
        Contrastive    & S    & \xmark     & 65.21           & 78.86           & CosFace        & L    & \xmark     & 82.72          & 96.65           \\
        Center         & S    & \xmark     & 65.49           & 80.14           & UniTsFace      & L    & \xmark     & \textbf{85.01} & \textbf{97.85}  \\
        L-Softmax      & S    & \xmark     & 67.12           & 80.42           & EPL            & L    & \xmark     & 83.89          & 97.44           \\ 
        \cline{6-10}
        CosFace        & S    & \xmark     & 77.11           & 89.88           & CosFace        & L    & \cmark    & 97.91          & 97.91           \\
        ArcFace        & S    & \xmark     & 77.50           & 92.34           & ArcFace        & L    & \cmark    & 98.35          & 98.48           \\
        UniFace(m=0.4) & S    & \xmark     & 77.31           & 92.56           & Circle Loss    & L    & \cmark    & 98.50          & 98.73           \\
        UniTSFace      & S    & \xmark     & 77.41           & \textbf{93.50 } & CurricularFace & L    & \cmark    & 98.71          & 98.64           \\
        EPL            & S    & \xmark     & \textbf{78.02 } & 93.15           & VPL            & L    & \cmark    & 98.80          & 98.97           \\ 
        \arrayrulecolor{black}\cline{1-2}\arrayrulecolor{black}\cline{3-5}
        ArcFace        & S    & \cmark     & 91.75           & 93.69           & Partial FC     & L    & \cmark    & 98.94          & 99.10           \\
        UniFace(m=0.4) & S    & \cmark     & 92.27           & \textbf{95.06 } & UNPG           & L    & \cmark    & 99.27          & -               \\
        UniTSFace      & S    & \cmark     & 92.36           & 94.84           & UniTsFace      & L    & \cmark    & 99.27          & \textbf{99.19}  \\
        EPL            & S    & \cmark     & \textbf{92.95 } & 94.50           & EPL            & L    & \cmark    & \textbf{99.31} & 98.79           \\
        \arrayrulecolor{black}\hline
        \end{tabular}
    }
\end{table*}

\subsection{Benchmark Results}
Using the proposed EPL, we train ResNet50 and ResNet100 on CASIA-WebFace0.5M, WebFace4M, WebFace12M, respectively, and compare the performance of EPL with the previous face recognition methods, including Contrastive Loss~\cite{chopra_learning_2005, hadsell_dimensionality_2006}, (N+1)-Tuplet Loss~\cite{schroff_facenet_2015}, CosFace~\cite{wang_cosface_2018}, ArcFace~\cite{deng_arcface_2022}, UniFace~\cite{zhou_uniface_2023}, VPL~\cite{deng_variational_2021}, UNPG~\cite{jung_unified_2022}, and the recent UniTSFace~\cite{li_2023_unitsface}, on two popular benchmarks.

\paragraph{MFR Ongoing Benchmarks}
Table \ref{tab:mfr_benchmark_results} reports the evaluation results on the MFR Ongoing benchmark, 
we can find that EPL consistently outperforms the other methods on the MR-All, Mask, and the cross-racial datasets. Specifically, when training a ResNet50 with the CASIA-WebFace0.5M, EPL achieves 40.92\% and 51.92\% scores on the Mask and MR-All, which are 2.94\% and 1.64\% higher than those of UniTSFace (37.98\% and 50.28\%), respectively. On IJB-C, the performance (83.38\% and 90.13\% for TAR@FAR=1e-5 and 1e-4) of EPL is also higher than that (82.64\% and 89.84\%) of UniTSFace. Moreover, when using large datasets such as WebFace4M and WebFace12M, EPL achieves higher performance on MR-All (89.76\% and 93.14\%) than other competing methods. Additionally, EPL's performance  (95.73\%) is still optimal when using a large backbone such as ResNet100 and training it with WebFace12M. These results show the effectiveness of our EPL. 

Our EPL achieved significant improvements on challenging datasets such as Mask, MR-All, and IJB-C datasets. However, on the subsets including LFW, CFP, and Age, the performance of EPL is comparable to UniTSFace. We believe this is in line with our expectation as 1) the amount of images in such datasets is far less than the Mask, MR-All, and IJB-C datasets and 2) the performance on these small datasets can have large fluctuations even for a few wrongly predicted samples.

\paragraph{MegaFace Challenge 1}
Table~\ref{tab:megaface_result} compares the identification and verification performance of the proposed EPL with the previous state-of-the-art methods on the MegaFace Challenge 1.
For a fair comparison, we follow the official protocols and evaluate the proposed EPL, trained on the CASIA-WebFace dataset, against models trained on `Small' datasets. Additionally, we compared EPL, trained on the Glint360K dataset, with models trained on `Large' datasets.

With the `small' protocol, our EPL achieved the highest accuracy (92.95\% and 78.02\%) on the identification track regardless of whether label refinement is used or not.
With the `large' protocol, our EPL achieved the optimal performance (99.31\%) on the identification track when labal refinement is used. When label refinement is not used, our EPL lower than UniTSFace. 
On the verification track, EPL's performance is comparable with UniTSFace. 
Because our EPL pulls hard samples via construct the EP and adaptive margin. EPL assigns larger margins to the normal samples and smaller margins to the hard samples during the model training. Therefore, our EPL is still a prototype learning method, which is beneficial for $1:n$ identification task. In contrast, UniTSFace discriminate positive sample pairs and negative sample pairs via learnable unified threshold on instance-wise, which is beneficial for $1:1$ verification task.

\section{Conclusion}
In this paper, we define the empirical prototype and present the empirical prototype learning (EPL),
which can be combined with the existing prototype learning based methods, and consistently improve their performance in the face recognition. EPL performs well on the MFR Ongoing Challenge, outperforming previous state-of-the-art method in the $1\text{:}n$ face identification. However, due to that EPL is a prototype based method in nature, its performance in the $1\text{:}1$ face verification is slightly lower and requires further improvement.

\bibliographystyle{elsarticle-num}
\bibliography{references}

\end{document}